\documentclass{article}

\usepackage[dblblindworkshop, final]{neurips_2025}
\workshoptitle{The 1st Workshop on
Efficient Reasoning}

\usepackage[utf8]{inputenc} 
\usepackage[T1]{fontenc}    
\usepackage{hyperref}       
\usepackage{url}            
\usepackage{booktabs}       
\usepackage{amsfonts}       
\usepackage{nicefrac}       
\usepackage{microtype}      
\usepackage{xcolor}         

\usepackage{subcaption}
\usepackage{tcolorbox}

\usepackage{multirow}
\usepackage{graphicx}
\usepackage{subcaption}
\usepackage{enumitem} 
\usepackage{amsmath}
\usepackage{amssymb}
\setlist[itemize]{leftmargin=5mm}
\usepackage{amsmath, amssymb, mathtools, nccmath}

\usepackage{adjustbox}
\usepackage{float}
\usepackage{tabularray}
\usepackage{tabularx}
\usepackage{sidecap}
\usepackage{comment}
\usepackage{wrapfig}
\usepackage{booktabs,tabularx, colortbl} 
\usepackage{lipsum}
\usepackage[dvipsnames]{xcolor}
\usepackage{amsmath, amsthm, amssymb}
\usepackage{graphicx}
\usepackage{hyperref}
\usepackage{cleveref}
\usepackage{soul}
\usepackage{}

\usepackage{booktabs}    
\usepackage{makecell}    
\usepackage{pifont}      

\usepackage{subcaption} 
\usepackage{tabularx}
\usepackage{booktabs}
\usepackage{siunitx}
\usepackage{caption}
\title{SRT: Accelerating Reinforcement Learning via Speculative Rollout with Tree-Structured Cache}

\author{
Chi-Chih Chang$^1$\thanks{Equal contribution. Work during internship at Bytedance.} \quad Siqi Zhu$^2$\footnotemark[1] \quad  Zhichen Zeng$^4$\quad Haibin Lin$^5$ \\
\bf Jiaxuan You$^2$ \quad Mohamed S. Abdelfattah$^1$ \quad Ziheng Jiang$^5$\quad Xuehai Qian$^3$\\
$^1$Cornell University \quad $^2$University of Illinois Urbana-Champaign \quad \\
$^3$Tsinghua University \quad $^4$University of Washington \quad $^5$ByteDance
}

\begin{document}

\maketitle

\begin{abstract}
We present Speculative Rollout with Tree-Structured Cache (SRT), a simple, model-free approach to accelerate on-policy reinforcement learning (RL) for language models without sacrificing distributional correctness. SRT exploits the empirical similarity of rollouts for the same prompt across training steps by storing previously generated continuations in a per-prompt tree-structured cache. During generation, the current policy uses this tree as the draft model for performing speculative decoding. To keep the cache fresh and improve draft model quality, SRT updates trees online from ongoing rollouts and proactively performs run-ahead generation during idle GPU bubbles. Integrated into standard RL pipelines (\textit{e.g.}, PPO, GRPO and DAPO) and multi-turn settings, SRT consistently reduces generation and step latency and lowers per-token inference cost, achieving up to 2.08× wall-clock time speedup during rollout. 
\end{abstract}

\section{Introduction}
Reinforcement learning (RL) has emerged as a pivotal paradigm for scaling language models, enabling them to tackle sophisticated problems like competition-level mathematics and programming tasks through deeper and longer reasoning processes~\citep{5team2025glm45, deepseekai2025deepseekr1, 2025seed15thinking, openai2025gptoss120b, yang2025qwen3, kimiteam2025kimi}.

\vspace{-0.2cm}
\begin{wrapfigure}[16]{hr}{0.5\textwidth}
\begin{center}
\vspace{-11pt}
    \includegraphics[width=0.44\textwidth, trim = 0 0.8cm 0 1.1cm]{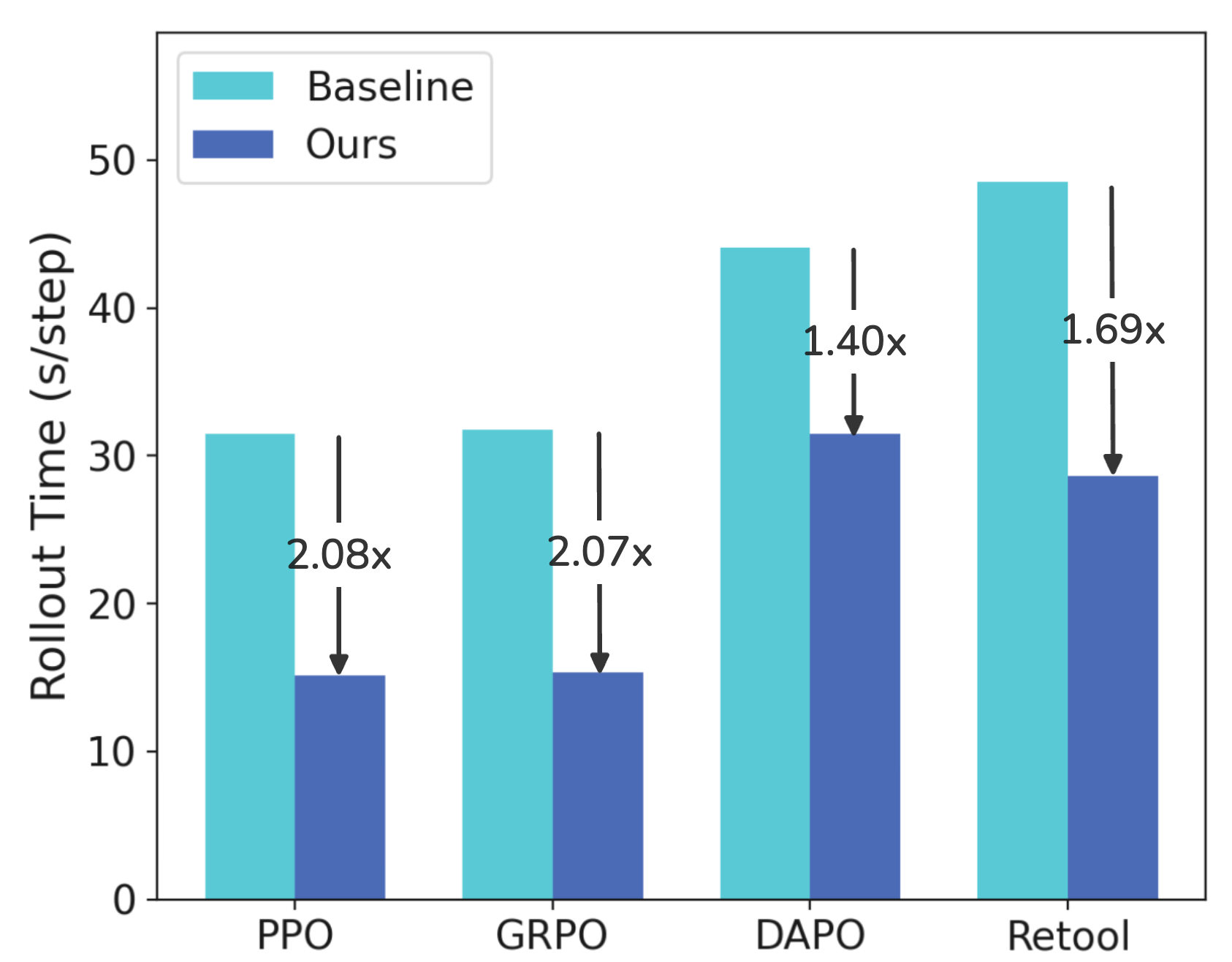}
\end{center}
\caption{Rollout Speedups of SRT Across Different RL Algorithms on Qwen2.5-1.5B}
\label{fig: teaser}
\vspace{-0.5cm}
\end{wrapfigure}
\begin{figure}[ht]
    \centering
    \begin{subfigure}[t]{0.32\textwidth}
        \centering
        \includegraphics[width=\linewidth]{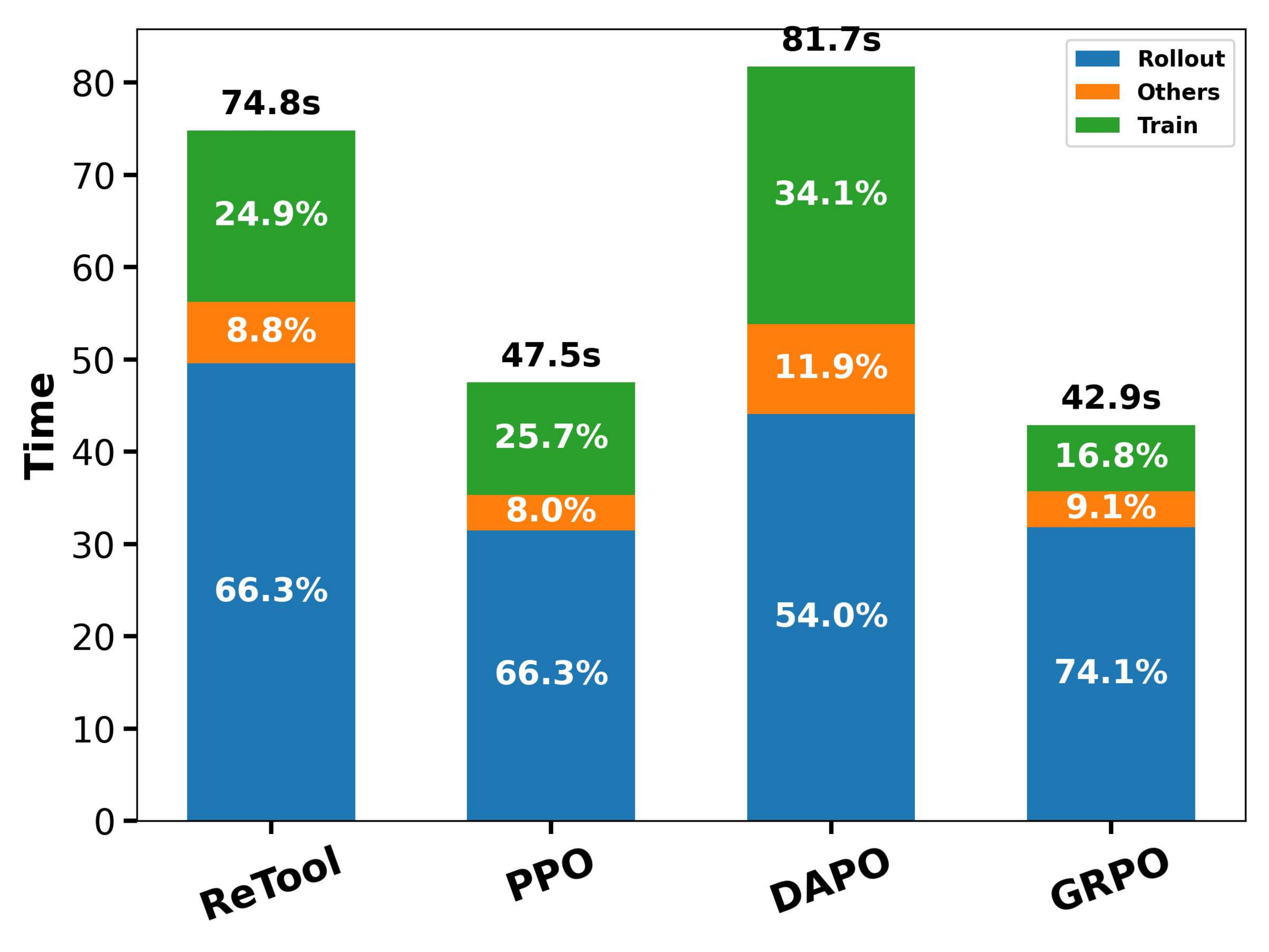}
        \label{fig:table1}
    \end{subfigure}\hfill
    \begin{subfigure}[t]{0.32\textwidth}
        \centering
        \includegraphics[width=\linewidth]{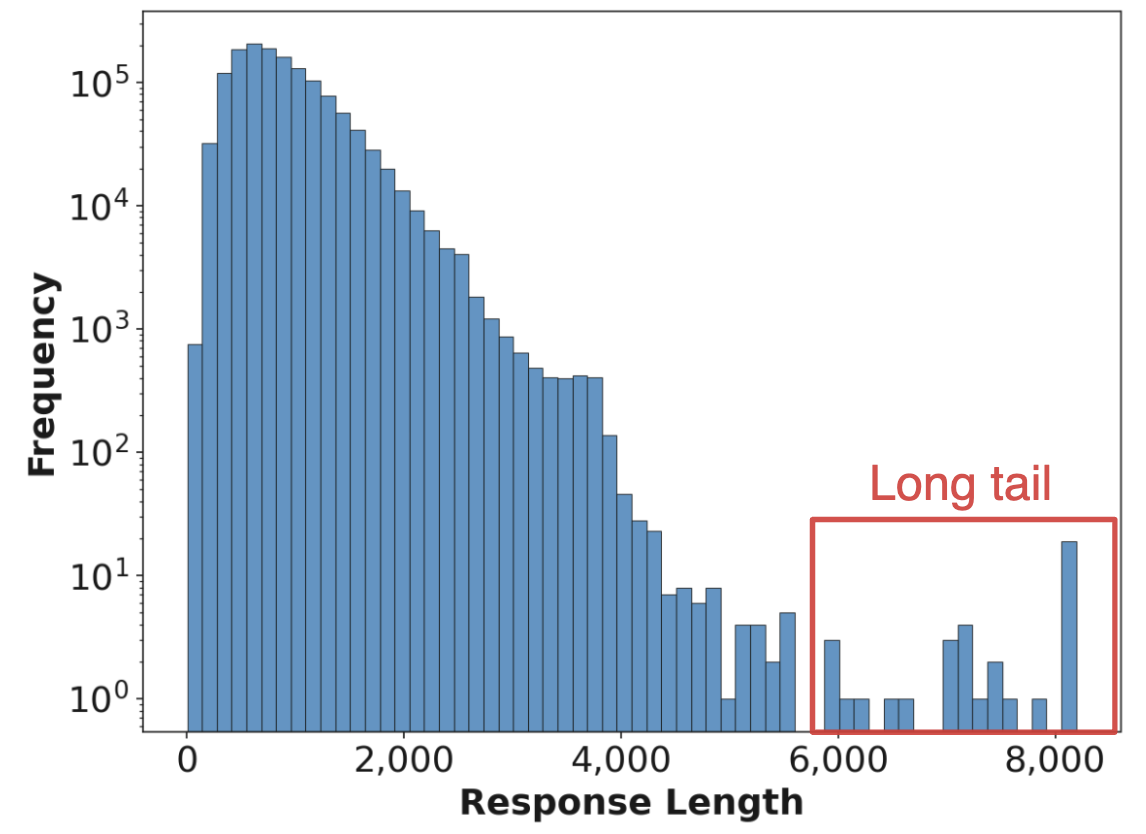}
        \label{fig:output_length}
    \end{subfigure}\hfill
     \begin{subfigure}[t]{0.32\textwidth}
        \centering
        \includegraphics[width=\linewidth]{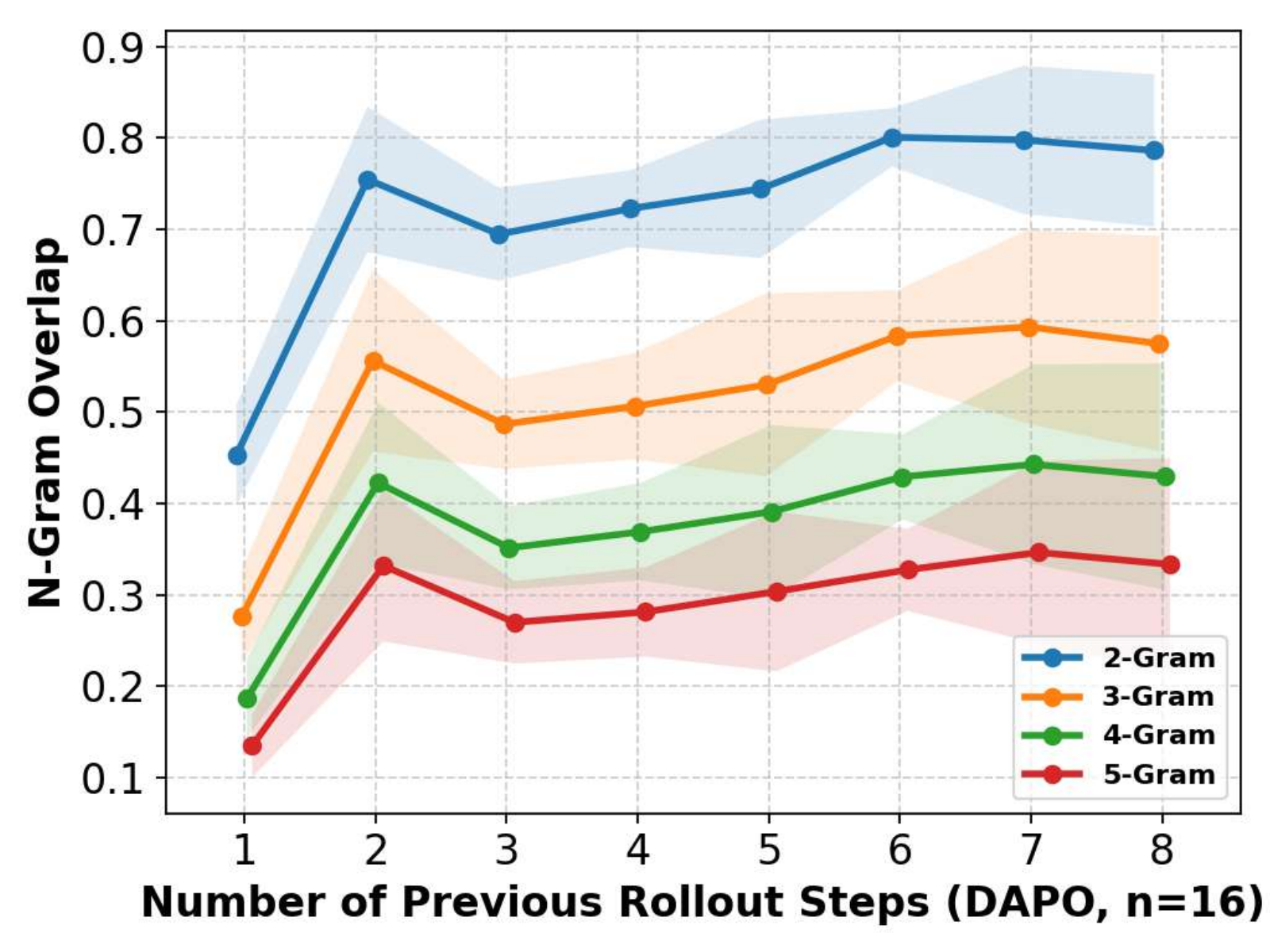}
        \label{fig:ngram}
    \end{subfigure}
    \vspace{-0.5cm}
    \caption[Caption for xxx]{\textbf{(a)}. Time breakdown across different RL algorithm.  \textbf{(b.)} Output length distribution on DAPO-17k dataset.
    \textbf{(c.)} Example of N-gram overlap for a specific prompt. The overlap is computed by comparing rollouts from the current step against the aggregated N-grams from all prior steps.\protect\footnotemark}
    \label{fig:all_three}
\end{figure}

However, as RL training scales, the rollout generation phase has become the dominant wall-clock bottleneck, consuming over 70\% (see Figure ~\ref{fig:all_three}) of total runtime in synchronous systems~\citep{Sheng25verl}. This stems from the auto-regressive, memory-bound nature of token generation and the ``long-tail'' distribution of response lengths, where a few lengthy rollouts stall entire batches, leaving GPUs idle and underutilized~\citep{fu2025areal, zhong2025rlhffuse}. The issue is exacerbated in advanced algorithms like Group Relative Policy Optimization (GRPO)~\citep{shao2024grpo} and Decoupled Clip and Dynamic Sampling Policy Optimization (DAPO)~\citep{yu2025dapo}, which sample multiple responses per prompt to compute relative advantages or filter uninformative data, thereby amplifying generation costs.

To mitigate rollout overhead, recent algorithm-system co-design approaches~\citep{fu2025areal, 2025seed15thinking, kimiteam2025kimi} relax strict on-policy execution by asynchronously launching future rollouts without awaiting current weight updates. While these approaches boost hardware utilization and throughput by reducing idle ``bubbles'', they deviate from on-policy sampling, potentially introducing convergence issues in certain regimes. 

In this work, we introduce \textit{Speculative Rollout with Tree-Structured Cache} (SRT), a new acceleration paradigm that speeds up on-policy rollouts without sacrificing training
efficiency.
SRT leverages the observation that 
the policies at different training epochs
exhibit a certain level of similarity when
responding to the same prompt. 
We maintain the generated rollouts of questions and store them in a tree-structured cache, using them as a model-free draft for speculative decoding during generation: the current policy verifies and accepts drafted tokens up to the first mismatch, ensuring lossless preservation of the on-policy distribution. 

To improve cache freshness and draft quality, SRT employs two complementary cache maintenance strategies.
First, it inserts decoded tokens from ongoing rollouts into the tree-structured cache.
Second, near the end of each step—when only a few long sequences remain—SRT leverages otherwise idle GPUs to run ahead on active prompts, generating partial rollouts that are inserted into the cache and then discarded.
This preserves on-policy training: actual rollouts are produced at the designated step, and speculative continuations are never used as learning targets.




We evaluate SRT on popular RL training algorithms (PPO, GRPO, DAPO) and in the multi-turn scenario (ReTool~\cite{feng2025retoolreinforcementlearningstrategic}). Extensive experiments show that SRT can achieve rollout speedups of up to 2.08$\times$ providing a practical, on-policy path to more efficient RL training.

\vspace{-0.2cm}

\footnotetext{Experiments are conducted using Qwen2.5-1.5B}

\section{Motivation}

\paragraph{Dominant rollout generation time during RL training.}The standard RL pipeline suffers from substantial inefficiencies, with the rollout stage emerging as the dominant computational bottleneck. As shown in Figure~\ref{fig:all_three}a, rollout consumes 65\% of total training time across four algorithms on average, making it the most critical target for optimization. 

\paragraph{Imbalanced response length among
questions in a batch.} A further structural issue arises from batched generation. Output lengths follow a long-tailed distribution (Figure~\ref{fig:all_three}b): while most sequences terminate quickly, a few very long responses delay batch completion. This imbalance leaves many GPUs idle, creating underutilized bubbles of compute.

\paragraph{Similarity of rollouts across epochs.} 
Responses to the same question often exhibit high similarity across epochs, and this overlap grows as more rollouts are accumulated (Figure~\ref{fig:all_three}c). 
This pattern suggests an opportunity
that existing RL methods fail to leverage: 
historical responses can be used to 
predict current rollouts to accelerate 
training. This missed opportunity wastes GPU resources on generating nearly identical content.



These factors point to a clear need for rethinking the standard RL pipeline and motivate the development of more streamlined RL strategies.


\vspace{-5pt}
\section{Method}

\paragraph{RL training in brief.}
We consider standard on-policy reinforcement learning for language models. Given a prompt $x_{1:m}$ and a policy $\pi_{\theta}$, the learner samples one or more continuations $y$ by autoregressively decoding from $\pi_{\theta}(\,\cdot \mid x_{1:m})$. A task-specific reward $r(x,y)$ is computed (e.g., program execution, self-consistency or preference modeling), and gradients are formed from advantages $A(x,y)$ under a clipped policy-gradient objective (PPO-style) or its multi-sample variants such as GRPO and DAPO. Training alternates between a \emph{rollout} phase that generates $K$ samples per prompt and an \emph{update} phase that fits $\pi_{\theta}$ on the on-policy batch.

\paragraph{Per‑prompt rollout cache using tree-structured cache.}
SRT accelerates these on-policy rollouts by maintaining, for each prompt $p$, a cache of previously seen token subsequences organized as a tree-structured cache $\mathcal{T}_p$. Paths in $\mathcal{T}_p$ therefore compactly index \emph{all} substrings that have occurred in earlier generations for the same prompt, including from prior policy checkpoints. Each node corresponds to a context (a token subsequence) and stores outgoing edges labeled by next tokens, together with simple frequency statistics: $\texttt{count}(u)$ for a node $u$ recording its frequency in previous rollouts. This structure is purely model-free and can be stored in CPU memory; it can be updated online in amortized linear time as new tokens arrive. 

\vspace{-0.2cm}
\begin{wrapfigure}[15]{hr}{0.47\textwidth}
\begin{center}
    \includegraphics[width=0.47\textwidth, trim = 0 0.8cm 0 1.3cm]{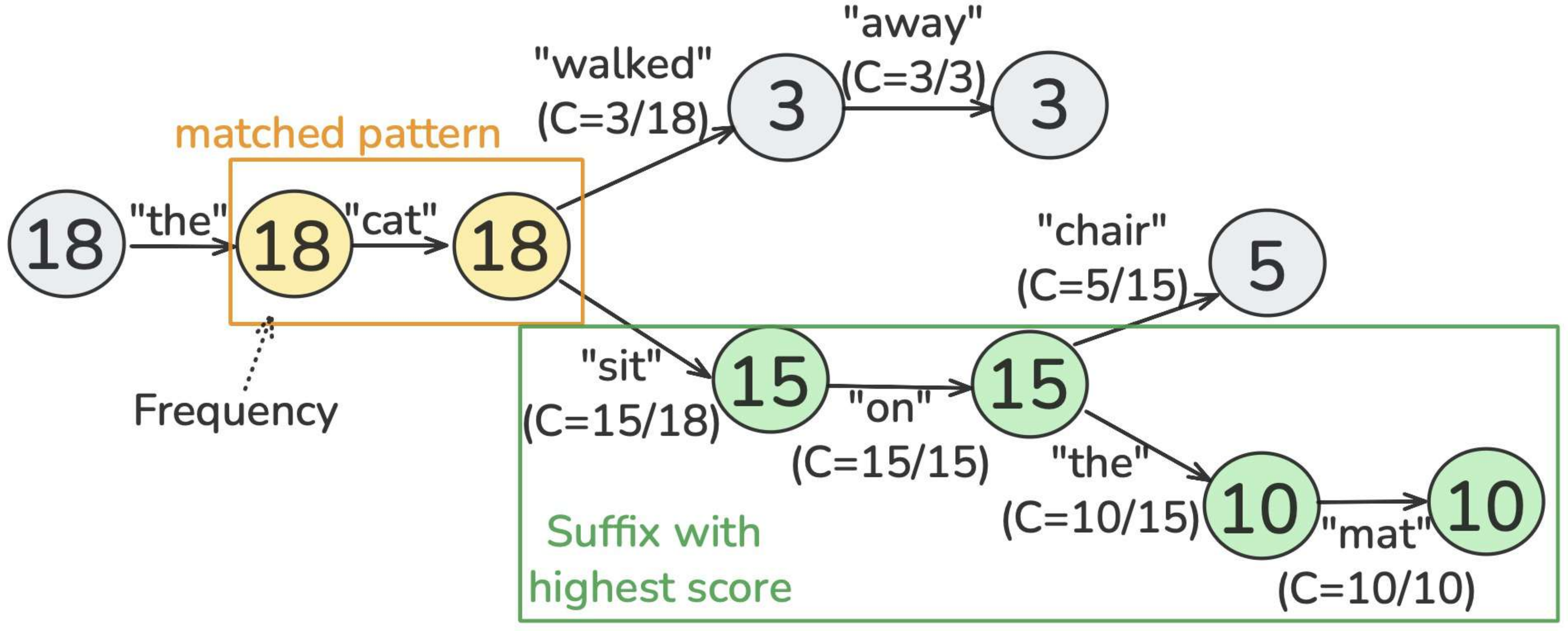}
\end{center}
\caption{Given the matched prefix ``the cat'', we choose its suffix ``sit on the mat'' with highest score as draft tokens. Leaf nodes show the number of times each suffix appeared in cached rollouts, while non-leaf nodes contain the sum of their children's counts.}
\label{fig: tree_example}
\vspace{-0.2cm}
\end{wrapfigure}

\paragraph{Speculative Rollout with Tree-Structured Cache.}
During rollout for prompt $p$, suppose we have already produced a partial continuation $y_{1:t}$. We locate the longest suffix of $y_{1:t}$ that appears in $\mathcal{T}_p$ by walking through the tree from the root along tokens $y_{t-q+1:t}$; if the walk fails, we revert to standard decoding for one step and try again. Once a match of length $q$ is found at node $u_q$, SRT assembles a set $\widehat{\mathcal{T}}$ of draft tokens by greedily adding descendants of $u_q$ that are most likely to be accepted. We rank an edge to child $v$ by empirical conditional
\[
C(v) \;=\; \frac{\texttt{count}(v)}{\sum_{w \in \texttt{children}(\texttt{parent}(v))} \texttt{count}(w)} ,
\]
and score a node by the product of scores along its path from $u_q$. Intuitively, $C(v)$ estimates how often a specific next token followed this prefix in prior rollouts, and the path product estimates the chance that a drafted chain of tokens will align with what the current policy would generate. We continue expansion until a budget $B(q)$ is reached. Given $\widehat{\mathcal{T}}$, SRT performs one decode pass of the current policy to verify multiple drafted tokens in parallel following classic speculative decoding procedure~\citep{cai2024medusasimplellminference,Miao2023SpecInfer,Li2024EAGLE,pmlr-v202-leviathan23a,oliaro2025suffixdecode}.

\vspace{-0.2cm}
\begin{figure}[h]
    \centering
    \includegraphics[width=1.0\linewidth]{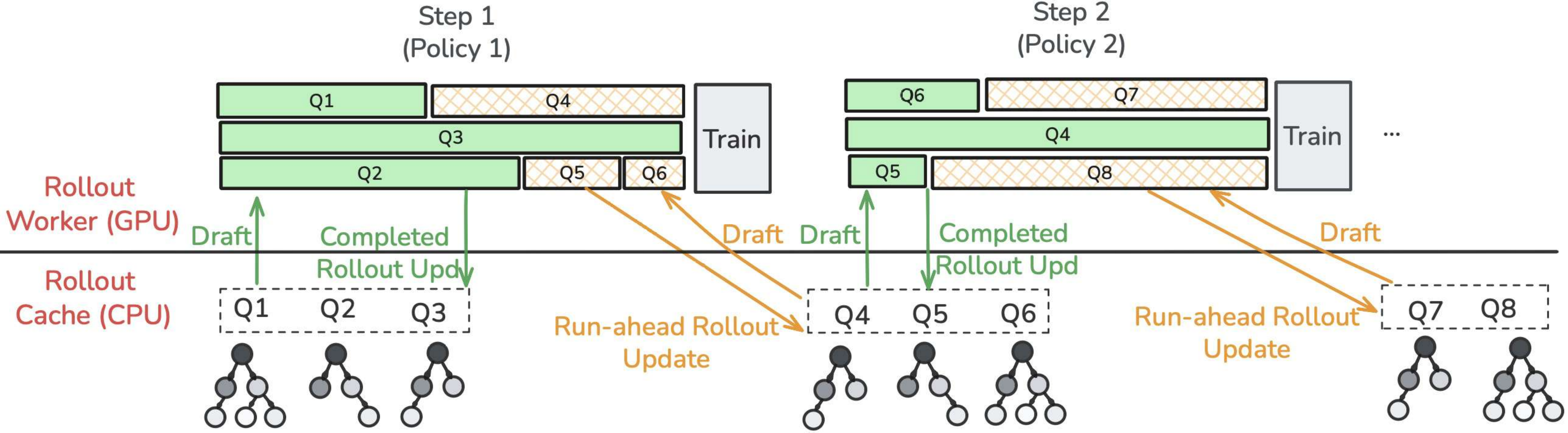}
    \caption{Illustration of cache maintenance strategy in SRT.}
    \label{fig:overview}
\end{figure}
\vspace{-0.5cm}

\paragraph{Cache update strategy.}
The maintenance of the rollout cache is crucial to the achievable speedups. In SRT, we maintain the cache by two sources as illustrated in Figure~\ref{fig:overview}. First, decoded outputs of \emph{running rollouts} are inserted online into $\mathcal{T}_p$ and node counts are updated. This immediately benefits the remaining samples for the same prompt in multi-sample algorithms and carries signal across training steps when the prompt reappears. Second, we exploit \emph{run-ahead generation} during bubbles. Whenever some sequences in a batch finish early and GPU compute would have otherwise been idle, we allocate that slack to generate rollouts of prompts that will be sampled soon (e.g., from the data loader’s look‑ahead window or an active prompt queue). Run‑ahead tokens are inserted into $\mathcal{T}_p$ but are never used for learning targets; they serve solely as future drafting hints. Unlike a history-only caches design \citep{he2025rhymerl}, which caches only completed responses from previous epochs in an offline and asynchronous fashion, SRT's maintenance strategy mitigate cold-start for first-time prompts, actively enriches the cache, and yields more accepted tokens per decoding step—reducing decoding steps and end-to-end latency.


\vspace{-10pt}
\section{Experiment}

\subsection{End-to-end Results}

\begin{table}[ht]
  \centering
  \caption{\textbf{SRT achieves superior performance over other methods.}}
  \label{tab:e2e main results}
  \renewcommand{\arraystretch}{1.2}
 \resizebox{0.75\textwidth}{!}{
  \begin{tabularx}{0.8\textwidth}{l *{12}{>{\centering\arraybackslash}X}}
    \toprule
    & \multicolumn{4}{c}{\textbf{Gen (s) (↓)}} 
    & \multicolumn{4}{c}{\textbf{Step (s) (↓)}} 
    & \multicolumn{4}{c}{\textbf{$\mu s$/token (↓)}} \\
    \cmidrule(lr){2-5} \cmidrule(lr){6-9} \cmidrule(lr){10-13}
    \textbf{Method} 
      & \rotatebox{30}{PPO} & \rotatebox{30}{GRPO} & \rotatebox{30}{DAPO} & \rotatebox{30}{Retool}
      & \rotatebox{30}{PPO} & \rotatebox{30}{GRPO} & \rotatebox{30}{DAPO} & \rotatebox{30}{Retool}
      & \rotatebox{30}{PPO} & \rotatebox{30}{GRPO} & \rotatebox{30}{DAPO} & \rotatebox{30}{Retool} \\
    \midrule
    Baseline          &  31.5 &  31.8 & 44.1  & 49.0  & 47.5  & 42.9  &  81.7 & 74.8  & 104  &  83.8 &  32.9 & 121  \\
    N-gram            & 31.4  & 31.1  & 46.0  & 45.0  & 47.3  & 42.1  &  84.5 & 69.3  & 105  & 82.4  &  33.3 & 161  \\
    SuffixDecoding     &  18.4 &  19.7 & 62.5  &  38.8 &  35.9 & 30.7  & 103  & 69.2  & 56.6  & 52.4  &  41.1 & 76.7  \\
    SRT (Ours) &  \textbf{15.2} &  \textbf{15.4} &  \textbf{31.5} & \textbf{28.7}  & \textbf{31.5}  & \textbf{26.2}  & \textbf{68.7}  &  \textbf{58.8} & \textbf{48.3}  & \textbf{41.6}  & \textbf{23.3}  & \textbf{62.2}  \\
    \bottomrule
  \end{tabularx}
  }
\end{table}




\paragraph{Effectiveness of Per-prompt Rollout Cache.} We present the experiment results without run-ahead generation to examine the potential of the proposed rollout cache. We integrated SRT into vLLM~\citep{kwon2023efficientmemorymanagementlarge} and used it as the inference engine for Verl~\citep{Sheng25verl}. Our end-to-end experiments were conducted with the Qwen2.5-1.5B model, using on-policy RL across four algorithms. Concretely, PPO and GRPO were trained on the math dataset~\citep{hendrycks2021measuringmathematicalproblemsolving}, while DAPO and ReTool were trained on DAPO-Math-17k~\citep{yu2025dapoopensourcellmreinforcement}. As summarized in Table~\ref{tab:e2e main results}, SRT consistently outperforms speculative decoding strategies that were originally designed for non-RL scenario across various RL algorithms. In particular, SRT achieves lower generation and step latency as well as reduced per-token inference cost, and these gains hold robustly across both single-turn (PPO/GRPO/DAPO) and multi-turn (ReTool) training regimes.


\subsection{Effect of On-the-Fly Updates and Run-Ahead Generation}

\begin{wrapfigure}[13]{hr}{0.3\textwidth}
\begin{center}
    \includegraphics[width=0.3\textwidth, trim = 0 0.8cm 0 4.8cm]{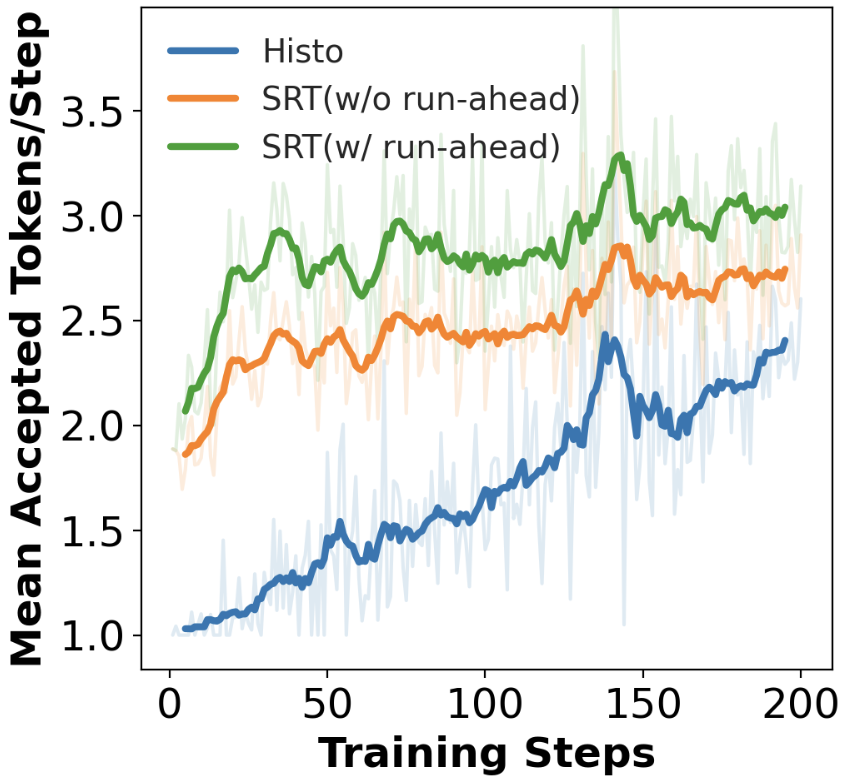}
\end{center}
\caption{Mean accepted tokens analysis of different cache maintenance strategy.}
\label{fig:runahead}
\end{wrapfigure}

In this section, we present simulation studies evaluating the impact of (i) on-the-fly updates from ongoing rollouts and (ii) run-ahead generation, both of which enrich the tree-structured cache. Using the DAPO algorithm on DAPO-17k~\citep{yu2025dapo}, we compare SRT to a history-only baseline that updates the cache solely with fully completed responses from previous epochs. As shown in Fig.~\ref{fig:runahead}, SRT consistently achieves a higher mean of accepted tokens (per decoding step). Enabling run-ahead yields additional gains, indicating that a richer cache produces higher-quality drafts. In practice, this reduces decoding steps per prompt, thereby reducing end-to-end cost and latency.
\section{Conclusion}

We introduced Speculative Rollout with Tree-Structured Cache (SRT), a speculative rollout algorithm that accelerates on-policy reinforcement learning by exploiting redundancy across rollouts. By organizing past continuations into per-prompt tree-structured caches and leveraging speculative decoding with run-ahead generation, SRT achieves substantial reductions in generation latency and inference cost without compromising distributional correctness. Experiments across multiple RL algorithms and multi-turn settings demonstrate up to 2.08× rollout speedup, highlighting SRT as a practical framework for scaling efficient RL training.

\newpage
\bibliographystyle{abbrvnat}
\bibliography{neurips_2025}


\newpage
\appendix
\section{Experimental Setup}

\begin{table}[h]
\centering
\caption{Training configurations for four algorithms.}
\label{tab:traincfg}

\begin{minipage}{0.48\textwidth}
\centering
\textbf{(a) PPO}\\[3pt]
\resizebox{\textwidth}{!}{
\begin{tabular}{lclc}
\hline
\textbf{Hyperparameter} & \textbf{Value} & \textbf{Hyperparameter} & \textbf{Value} \\
\hline
Actor learning rate     & $1 \times 10^{-6}$ & Critic learning rate    & $1 \times 10^{-5}$ \\
Warmup ratio            & 0.0                & Rollout temperature     & 1.0 \\
KL Coefficient ($\beta$) & 0.001              & Train batch size        & 512 \\
PPO mini batch size     & 128                & Training steps          & 300\\
Max input length        & 512      & Max response length     & 4096 \\
\hline
\end{tabular}}
\end{minipage}\hfill
\begin{minipage}{0.48\textwidth}
\centering
\textbf{(b) GRPO}\\[3pt]
\resizebox{\textwidth}{!}{
\begin{tabular}{lclc}
\hline
\textbf{Hyperparameter} & \textbf{Value} & \textbf{Hyperparameter} & \textbf{Value} \\
\hline
Actor learning rate     & $1 \times 10^{-6}$ & Response per Prompt & 5 \\
Warmup ratio            & 0.0                & Rollout temperature     & 1.0 \\
KL Coefficient ($\beta$) & 0.001              & Train batch size        & 128 \\
PPO mini batch size     & 64                & Training steps          & 300\\
Max input length        & 512      & Max response length     & 4096 \\
\hline
\end{tabular}}
\end{minipage}

\vspace{1em}

\begin{minipage}{0.48\textwidth}
\centering
\textbf{(c) DAPO}\\[3pt]
\resizebox{\textwidth}{!}{
\begin{tabular}{lclc}
\hline
\textbf{Hyperparameter} & \textbf{Value} & \textbf{Hyperparameter} & \textbf{Value} \\
\hline
Actor learning rate     & $1 \times 10^{-6}$ & Response per Prompt & 16 \\
Warmup ratio            & 0.0                & Rollout temperature     & 1.0 \\
KL Coefficient ($\beta$) & 0.0              & Train batch size        & 128 \\
PPO mini batch size     & 32                & Training steps          & 300\\
Max input length        & 2048      & Max response length     & 8192 \\
Clip ratio high        & 0.28      & Clip ratio low     & 0.20 \\
\hline
\end{tabular}}
\end{minipage}\hfill
\begin{minipage}{0.48\textwidth}
\centering
\textbf{(d) ReTool}\\[3pt]
\resizebox{\textwidth}{!}{
\begin{tabular}{lclc}
\hline
\textbf{Hyperparameter} & \textbf{Value} & \textbf{Hyperparameter} & \textbf{Value} \\
\hline
Actor learning rate     & $1 \times 10^{-6}$ & Critic learning rate     & $2 \times 10^{-6}$ \\
Warmup ratio            & 0.0                & Rollout temperature     & 1.0 \\
KL Coefficient ($\beta$) & 0.001              & Train batch size        & 512 \\
PPO mini batch size     & 128                & Training steps          & 300\\
Max input length        & 2048      & Max response length     & 16384 \\
Max turns        & 8      & Clip ratio high    & 0.28 \\
Clip ratio low    & 0.20 & & \\
\hline
\end{tabular}}
\end{minipage}

\end{table}

RL training is conducted using Verl~\citep{Sheng25verl}, with vLLM~\citep{kwon2023efficientmemorymanagementlarge} serving as the inference engine. The experimental configuration is summarized in Table~\ref{tab:traincfg}. All RL training experiments are performed on 8$\times$~NVIDIA Hopper GPUs.


\section{Ablation Study}
\label{ablation}

\subsection{Responses per Prompt in GRPO}

We analyzed the effect of varying the number of responses per prompt on the efficiency of SRT. As reported in Table~\ref{tab:grpo-n-compare}, increasing $n$ from 5 to 10 produces a greater performance improvement relative to competing algorithms. This behavior is consistent with expectation: as additional rollouts are incorporated, the resulting outputs exhibit greater similarity, thereby enabling SRT to exploit more reliable historical information during speculative decoding.

\begin{table}[ht]
  \centering
  \caption{Comparison of step time and generation time for GRPO.}
  \label{tab:grpo-n-compare}
  \small
  \begin{tabular}{
    l
    S[table-format=1.3] S[table-format=1.3]
    S[table-format=1.3] S[table-format=1.3]
  }
    \toprule
    & \multicolumn{2}{c}{\textbf{GRPO $n=5$}} & \multicolumn{2}{c}{\textbf{GRPO $n=10$}} \\
    \cmidrule(lr){2-3}\cmidrule(lr){4-5}
    \textbf{Method} & {\textbf{Step} (s)} & {\textbf{Gen} (s)} & {\textbf{Step}} & {\textbf{Gen} (s)} \\
    \midrule
    Baseline          &  {42.9} & {31.8} & {61.2} & {47.1} \\
    N-gram            &  {42.1} & {31.1} & {58.8} & {45.3} \\
    SuffixDecoding     &  {30.7} & {19.7} & {51.9} & {30.5} \\
    SRT &  {\textbf{26.2}} & {\textbf{15.4}} & {\textbf{41.2}} & {\textbf{20.4}} \\
    \midrule
    Improvement (\%) &  {14.7} & {21.8} & {20.6} & {33.1} \\
    \bottomrule
  \end{tabular}
  \vspace{2pt}
  \captionsetup{font=small}
\end{table}

\section{Related Work}
\label{related_work}

\textbf{Efficient Reinforcement Learning Frameworks for LLMs.} 
Recent open-source RL frameworks have democratized RL training~\citep{Sheng25verl,fu2025areallargescaleasynchronousreinforcement,hu2025openrlhfeasytousescalablehighperformance}. 
A common design choice is to employ Ray~\citep{moritz2018raydistributedframeworkemerging} to coordinate inference engines (e.g., vLLM~\citep{kwon2023efficientmemorymanagementlarge}, SGLang~\citep{zheng2024sglangefficientexecutionstructured}) with training engines (e.g., Megatron~\citep{shoeybi2020megatronlmtrainingmultibillionparameter}, FSDP~\citep{zhao2023pytorchfsdpexperiencesscaling}). 
Building on this foundation, researchers have proposed various approaches to further improve efficiency. 
For example, GRESO~\citep{zheng2025actpaysefficientreinforcement} accelerates training by skipping low-quality rollouts. Our closest concurrent work, RhymeRL~\citep{he2025historyrhymesacceleratingllm}, explore the speculative rollouts by caching \emph{historical} responses and updating its cache \emph{offline} with outputs from the previous epoch asynchronously, without modifying the cache during the current batch's rollout, which can face history-scarcity cold-start. In SRT, instead of ingesting completed responses from the previous epoch, we (i) stream tokens \emph{on-the-fly} from the \emph{current batch’s} rollouts into a per-prompt tree-structured cache, and (ii) perform run-ahead generation for the \emph{future} batch to proactively enrich the cache. This unique design mitigates RhymeRL's cold-start issue by enriching useful drafts immediately and continuously, increasing accepted tokens per step and reducing decoding cost, offering a complementary acceleration path for on-policy RL.




\textbf{Speculative Decoding.} Conventional LLM decoding requires accessing the KV cache at every step, making the process memory-bandwidth bound and limiting GPU efficiency. Speculative decoding mitigates this bottleneck by generating multiple candidate tokens in advance, either with a smaller draft model~\citep{pmlr-v202-leviathan23a,cai2024medusasimplellminference,Miao2023SpecInfer,Li2024EAGLE} or through retrieval-based strategies~\citep{oliaro2025suffixdecode,yang2023inferencereferencelosslessacceleration}. These speculative tokens are subsequently verified by the base model, reducing KV cache accesses to a single pass while increasing computational intensity. Building upon this foundation, our work extends speculative decoding by reusing previous rollouts, thereby further improving the efficiency of RL training.


\end{document}